\documentclass[10pt,twocolumn,twoside]{IEEEtran}

\usepackage[utf8]{inputenc}
 
\usepackage{amsthm}
\usepackage{amsbsy,amsmath,amscd,amssymb} 
\usepackage{bm}
\usepackage{graphicx}
\usepackage{lineno,hyperref}
\usepackage{xcolor}

\newcommand{\EE}{\mathbb{E}}
\newcommand{\GG}{\mathcal{G}}
\newcommand{\Var}{\operatorname{Var}}

\newcommand{\s}{\mathbf{s}}
\newcommand{\K}{\mathbf{K}}
\renewcommand{\P}{\mathbf{P}}
\newcommand{\J}{\mathbf{J}}
\renewcommand{\v}{\mathbf{v}}

\newtheorem{theorem}{Theorem}
\newtheorem{proposition}{Proposition}
\newtheorem{corollary}{Corollary}

\begin{document}

\title{Non asymptotic bounds in asynchronous sum-weight gossip protocols}

%
%
%

\author{David~Picard,
		Jérôme Fellus,
        Stéphane Garnier
\thanks{D. Picard is with LIGM, Ecole des Ponts, Univ. Gustave Eiffel, CNRS, Marne-la-Vallée, France ; S. Garnier is with ETIS, UMR 8051, Université Paris Seine, Université Cergy-Pontoise, ENSEA, CNRS, FRANCE, email: david.picard@enpc.fr, jfellus@gmail.com.}}

\maketitle

\begin{abstract}
This paper focuses on non-asymptotic diffusion time in asynchronous gossip protocols. Asynchronous gossip protocols are designed to perform distributed computation in a network of nodes by randomly exchanging messages on the associated graph. To achieve consensus among nodes, a minimal number of messages has to be exchanged. We provides a probabilistic bound to such number for the general case. We provide a explicit formula for fully connected graphs depending only on the number of nodes and an approximation for any graph depending on the spectrum of the graph.

\end{abstract}


\section{Introduction}

With the advent of machine learning and its massive needs in computation, a lot of attention has been recently put on distributed estimation~\cite{dimakis10ieee}.
In this context, a network of nodes, each node containing a subset of observations, has to solve a joint estimation problem by exchanging information (\emph{e.g.}, sensor networks~\cite{iutzeler12icassp, iutzeler13tsp}).
The challenges are three-fold: First, solving the distributed joint estimation problem can have increased computational complexity compared to the equivalent centralized problem. Second, we must take into account the intrinsic cost of exchanging information. Third, synchronous communications, where nodes wait for others to exchange data, can drastically slow down the optimization process.

To tackle that second issue, Randomized Gossip algorithms~\cite{gossipAveraging} have been successfully applied to a wide variety of machine learning problems like clustering~\cite{fellus13icdmw,difatta11icdmw}, dimensionality reduction~\cite{fellus15neurocomp}, kernel methods~\cite{colin} and deep neural networks~\cite{blot16nipsw}.
In randomized gossip algorithms, the estimation problem is cast as a consensus averaging problem. All nodes alternate between local computations that aim at solving the local problem and message exchanges that ensure consensus among nodes. Forcing consensus is required to ensure that all nodes converge to the solution of the joint problem.

In this paper, we are interested in the special case of random asynchronous gossip algorithms (also called push only)~\cite{kempe2003gossip, benezit10isit}. Notably, these algorithms do not involve any waiting since nodes do not expect an answer to their messages.
More precisely, we focus on a specific protocol where at each time step $t$, only one message is exchanged from a sender node $s$ to a receiver node $r$.
$s$ is drawn uniformly among all nodes, and $r$ is drawn uniformly from the neighbors of $s$.

We address the problem of the number of messages needed to ensure that all nodes received information from any given starting point with sufficient probability, which we call diffusion time. This number is very important since it gives a bound above which consensus becomes possible. As such, it can be used to set the minimal number of exchanged messages in push only algorithms where a sufficient consensus is required before proceeding to the next step. We provide several results in this respect: We give the expectation of the minimal number of message until full diffusion is reach that depends only on the link probability. We provide an upper bound of this number for the specific case of fully connected networks, which is similar to the ``telephone call problem'' in theorem 5.1 of~\cite{frieze1985shortest} and used to provide asymptotic bounds in~\cite{kempe2003gossip}. We also provide a larger upper bound for generic networks depending on the spectrum of the corresponding graph.

The organization of this paper is as follows: First we give a formal definition of the problem and propose our first result. Then, we study the specific case of fully connected graphs and provide a probabilistic upper bound to the minimal number of messages required to reach full diffusion. In section~\ref{sec:cheeger}, we give an upper bound for graphs with bottleneck based on Cheeger constant. Finally, in the last section, we show some simulation results to highlight the tightness of the bound for fully connected graphs as well as the influence of strong bottleneck on the diffusion time graph.

\section{Related work}
For the sake of completeness, we reproduce in this section several results from the literature about the convergence of randomized Gossip protocols.
To improve the readability with respect to notations and properties, we first start analyzing deterministic  communication patterns before considering randomized communications. Some of the proofs we provide are different from what is found in the original papers.

\subsection{Deterministic communications}
In most cases, distributed estimation problems can be cast as the computation of a distributed average that are easily solved using Gossip algorithms.
Specifically, we consider here a network of $N$ nodes, each hosting a value $x_i$ as well as a weight $w_i, 1 \leq i \leq N$.
We are interested in computing the weighted average:
\begin{align*}
\bar{x} =  \frac{\sum_{i=1}^N w_i x_i}{\sum_{i=1}^N w_i}
\end{align*}
In Gossip approaches, this weighted average is computed by performing communications among nodes to update the local estimate of the average.
Each node estimates $\sum_{i=1}^{N}w_i x_i$ using a local estimate $s_i$ and $\sum_{i=1}^N w_i$ using another local variable $\omega_i$.
The weighted average is then locally estimated by computing the ratio of $s_i$ and $\omega_i$.
We note $\s(t) = [s_1(t), \ldots, s_N(t)]^\top$ the vector concatenating all local estimates $s_i(t), 1\leq i \leq N$ at time $t$.
Similarly, we write $\bm{\omega}_i(t) = [\omega_1(t) \ldots, \omega_N(t)]^\top$ the vector concatenating all the weight estimates.

Communications can then be expressed in matrix form using a communication matrix $\K$:
\begin{align}
\label{eq:ite}
	\s(t+1)^\top = \s(t)^\top\K
\end{align}
Where the entries of $\K$ define how information is exchanged on the network.
In particular, we can see that any specific entry $s_i(t)$ is updated using a linear combination of all the possible estimates:
\begin{align*}
	s_i(t+1) = \sum_{j=1}^N \K_{ji} s_j(t)
\end{align*}
Each non-zero entry $\K_{ji}$ in $\K$ thus implies a message from $j$ to $i$ containing both its estimate $s_j(t)$ and a aggregation coefficient $\K_{ji}$.
In fact, it can be argued that $\K$ defines both the network structure (which nodes are connected together) and the communication protocol.

Several properties have to be enforced on $\K$ to ensure that iterating Equation \ref{eq:ite} leads to the correct weighted average.
Common assumptions are $\K \geq 0$ and $\K\mathbf{1} = \mathbf{1}$, meaning that $\K$ is a \emph{row stochastic} matrix.
This second assumption ($\mathbf{1}$ is a right eigenvector of $\K$) leads the well known mass-conservation property~\cite{benezit10isit}:
\begin{proposition}
	Let $\K$ be a row stochastic matrix, then the sum of local estimates is conserved:
	\begin{align*}
	\forall t, \s(t)^\top \mathbf{1} = \s(0)^\top \mathbf{1}
	\end{align*}
	\begin{IEEEproof}
	By recursion, we have
	\begin{align*}
	\s(t+1)^\top\mathbf{1} = \s(t)^\top\K\mathbf{1} = \s(t)^\top\mathbf{1} = \ldots = \s(0)^\top\mathbf{1}
	\end{align*}
	\end{IEEEproof}
\end{proposition}

In many symmetric gossip protocols, such as newscast~\cite{voulgaris2003robust}, a column stochasticity is also enforced on $\K$.
The column stochasticity can be expressed in terms of left eigenvector by $\mathbf{1}^\top\K = \mathbf{1}^\top$.
Having $\K$ being column stochastic allows for a stable average~\cite{benezit10isit}:
\begin{proposition}
	Let $\K$ be column stochastic ($\mathbf{1}^\top \K = \mathbf{1}$), if $\forall i, s_i(t) = \bar{x} $ then
	\begin{align*}
	\s(t+1)= \s(t)
	\end{align*}
	\begin{IEEEproof}
	We have $\s(t) = \bar{x}\mathbf{1}$, and thus $\s(t+1) = \s(t)^\top\K = \bar{x}\mathbf{1}^\top\K = \bar{x}\mathbf{1}^\top = \s(t)$
	\end{IEEEproof}
\end{proposition}

The problem of doubly stochastic matrices is that they lead to implicit synchronization.
Indeed, if a matrix is both row stochastic and column stochastic, then there are nodes that are both sending messages and receiving messages at the same time.
Such nodes have to wait for the expected messages before they proceed to the next iteration and as such many waits can be induced among the nodes.
To solve that issue, one way protocols have to be used, which means that the communication matrix can no longer be doubly stochastic.

In~\cite{kempe2003gossip}, the authors show that a row stochastic matrix is sufficient to perform the averaging.
In the original paper, the results are given for random matrices $\K(t)$, but they hold for deterministic matrices in the following.
Since the matrix is no longer column stochastic, it has a left eigenvector $\v$ different from $\mathbf{1}$ which has to be compensated for to keep the stability of the average.
This is achieved by considering a weight $\omega_i(t)$ associated with every estimate $s_i(t)$ that is updated using the same communication matrix $\K$.
Furthermore, in order to compute the average, we have to guarantee that each node will at some point have accumulated information from all other nodes.
In terms of graph properties, this corresponds to a strongly-connected aperiodic graph, and translate into $\K$ being irreducible and aperiodic (also termed \emph{primitive}), that is, $\exists m, \K^m > 0$.
The ratio of the estimate and the weights then converges to the desired average:
\begin{theorem}
\label{th:determ}
	Let $\K \geq 0$ be row-stochastic and primitive and $\s(0) = [w_0 x_0, \ldots, w_N x_N]$ and $\bm{\omega}(0) = [w_0, \ldots, w_N]$, then
	\begin{align*}
	\lim_{t \rightarrow +\infty} \frac{\s(t)^\top}{\bm{\omega}(t)^\top} = \bar{x}\bm{1}^\top
	\end{align*}
	With $\frac{a}{b}$ being the element wise division of $a$ by $b$.
	\begin{IEEEproof}
	Remark that
	\begin{align*}
	\lim_{t \rightarrow +\infty} \frac{\s(t)^\top}{\bm{\omega}(t)^\top} =\lim_{t \rightarrow +\infty} \frac{\s(0)^\top\K^t}{\bm{\omega}(0)^\top\K^t} 
	\end{align*}
	Since $\K$ is irreducible and row stochastic, we can use Perron-Frobenius theorem~\cite{meyer2000matrix} to show that:
	\begin{align*}
	\lim_{t \rightarrow +\infty} \K^t = \bm{1}\v^\top
	\end{align*}
	with $\v>\bm{0}$. Thus, 
	\begin{align*}
		\lim_{t \rightarrow +\infty} \s(0)\K^t = \left(\sum_i w_i x_i\right)\v^\top
	\end{align*}
	and
	\begin{align*}
	\lim_{t \rightarrow +\infty} \bm{\omega}(0)\K^t = \left(\sum_i w_i \right)\v^\top
	\end{align*}
	And the well defined element wise ratio converges to $\bar{x}$.
	\end{IEEEproof}
\end{theorem}
In fact since $\K$ corresponds to a Markov chain with $N$ states, the convergence can be studied using the subsequent theory~\cite{sinclair1992improved}. It can be showed that the convergence is obtain at an exponential rate depending on the second eigenvalue of $\K$.

\subsection{Randomized Gossip}
However, having a connected deterministic communication matrix $\K$ enforces having a static network (no node leaving or entering the network) as well as total knowledge of the network at every node.
Based on these considerations, Gossip algorithms use random communication matrices $\K(t)$ and are known as Randomized Gossip Protocols to highlight this fact.
The update equations become:
\begin{align*}
	\s(t+1)^\top &= \s(t)^\top\K(t+1)\\
	\bm{\omega}(t+1)^\top &= \bm{\omega}(t)^\top\K(t+1)
\end{align*}
Where the $\K(t)$ are independently sampled from a stationary distribution of random matrices such that $\EE[\K(t)] = \K$, with $\K$ primitive.
The convergence of these update rules to the desired average depends on the properties of the product of all $\K(t)$.
In fact, to use the same scheme that proves Theorem~\ref{th:determ}, we would need the following property, which is called strong ergodicity~\cite{seneta2006non,dobrushin1956central}:

\begin{align*}
	\exists \v, 	\lim_{t \rightarrow +\infty} \prod_{\tau=1}^t \K(\tau) = \bm{1}\v^\top
\end{align*}
This is typically the case when all $\K(t)$ have the same dominant left eigenvector $\v$. 
In particular, doubly stochastic matrices $\K(t)$ amounts to $\mathbf{v}=\mathbf{1}$.
As recalled in \cite{benezit10isit}, asynchronous Gossip do not allow doubly stochastic matrices, and thus only weak ergodicity properties can be obtained.
Weak ergodicity is equivalent to having a time varying left eigenvector for the product of $\K(t)$ \cite{dobrushin1956central}:
\begin{align}
	\exists \{\v(t)\}_{t > 0}, \lim_{t \rightarrow +\infty} \left( \prod_{\tau = 1}^t \K(\tau) - \bm{1}\v(t)^\top \right) = 0  \label{eq_weak_ergo}
\end{align}

Actually, weak ergodicity proves sufficient to ensure convergence of random gossip protocols:

\begin{theorem}[Bénézit \textit{et al}. \cite{benezit10isit}]
\label{th:random}
	Let $\{\K(t)\}$ be a sequence of stationary row stochastic matrices with weak ergodicity, $\s(0) = [w_0 x_0, \ldots, w_N x_N]$ and $\bm{\omega}(0) = [w_0, \ldots, w_N]$, then with probability 1:
	\begin{align*}
	\lim_{t \rightarrow +\infty} \frac{\s(t)^\top}{\bm{\omega}(t)^\top} = \bar{x}\bm{1}^\top
	\end{align*}
	With $\frac{a}{b}$ being the element wise division of $a$ by $b$.
	\begin{IEEEproof}
	Weak ergodicity means that there exists $\v(t)$ satisfying Equation \ref{eq_weak_ergo}. Thus we can write
	\begin{align*}
		&\lim_{t \rightarrow +\infty} \frac{\s(t)^\top}{\bm{\omega}(t)^\top} = \lim_{t \rightarrow +\infty} \frac{\s(0)^\top  \prod_{\tau = 1}^t\K(\tau)}{\bm{\omega}(0)^\top \prod_{\tau = 1}^t\K(\tau)}  \\
	&\qquad  = \lim_{t \rightarrow +\infty} \frac{\s(0)^\top \bm{1}\v(t)^\top}{\bm\omega(0)^\top\bm{1}\v(t)^\top} = \bar x\lim_{t \rightarrow +\infty} \frac{\v(t)^\top}{\v(t)^\top} = \bar x \bm{1}^\top
	\end{align*}
	Note that the last simplification is valid only because $\lim_{t\rightarrow \infty} \Pr[\v(t)>\bm{0}] =1$. This can be checked using the counterpart of Borel-Cantelli lemma and recalling the primivity of $\K$ and the independence of the $\K(t)$. Indeed, 
	 \begin{align*}
	&\exists m, \K^m >0 \ \Rightarrow\  \forall t>0, \mathbb{E}\left[\prod_{\tau = t}^{t+m}\K(\tau)\right] = \epsilon > 0 \ \Rightarrow \\ 
	& \forall t\geq m, \Pr[\v(t)>0]=p>0 \
	\Rightarrow \ 	\sum_t Pr[\v(t) > 0] = \infty
	\end{align*}
which is a necessary and sufficient condition for $\Pr[\operatorname{limsup}_{n\rightarrow+\infty} \v(t) > 0] = 1$. That is, $\forall i,j, \v(t) > 0$ appears infinitely often with probability 1.
	\end{IEEEproof}
\end{theorem}

Moreover, we can derive a explicit convergence bound, introducing Dobrushin's coefficient of ergodicity~\cite{dobrushin1956central}. The Dobrushin's coefficient of some matrix $\bm{A}$ is
\begin{align*}
\mu_S(\bm{A}) = \frac 12 \max_{ij}\sum_k{\left| \bm{A}_{ij} - \bm{A}_{kj} \right|}
\end{align*}

\begin{theorem}
Noting $\P(t) = \prod_{\tau=0}^t \K(t)$ and $\mu_S(\P(t))$ the Dobrushin coefficient of $\P(t)$, the following bound holds :
\begin{align}
\left\|\frac{\s(t)^\top}{\bm{\omega}(t)^\top} - \bar{x}\bm{1}\right\| \leq \frac{4\mu_S(\P(t))}{\min_{ij}\P_{ji}(t)}\|\bar{x}\bm{1}\|
\end{align}
, 
\label{th:random}
	\begin{IEEEproof}For any i, we have
	\begin{align*}
	\left\vert \frac{s_i(t)}{{\omega}_i(t)} - \bar{x} \right\vert &= \left\vert \frac{\sum_j \P_{ji}(t) s_j(0)}{\sum_j \P_{ji}(t)\omega_j(0)} - \frac{\sum_j s_j(0)}{\sum_j{\omega_j(0)}}\right\vert \\
	&\leq \vert\bar{x}\vert \left\vert \frac{\max_j\P_{ji}(t)}{\min_j \P_{ji}(t)} - 1\right\vert \\
	& = \vert\bar{x}\vert \left\vert \frac{\max_j\P_{ji}(t) - \min_j \P_{ji}(t)}{\min_j \P_{ji}(t)} \right\vert \\
	& = \vert\bar{x}\vert  \frac{\max_{jk} \left\vert\P_{ji}(t) - \P_{ki}(t)\right\vert}{\min_j \P_{ji}(t)} 
\end{align*}
By summing over $i$, we have
\begin{align*}
	\left\| \frac{\s(t)^\top}{\bm{\omega}(t)^\top} - \bar{x}\bm{1}^\top\right\|^2 &\leq \frac{\sum_i \max_{jk} \left(\P_{ji}(t) - \P_{ki}(t)\right)^2}{\min_{ij} \P_{ji}(t)^2} \|\bar{x}\bm{1}\|^2
\end{align*}
Note $\mu = \frac{1}{N}\sum_k a_k$ and remark that
\begin{align*}
\max_{jk} (a_j - a_k)^2 &=& \max_{jk} (a_j - \mu + \mu - a_k)^2 \\
&=& \max_{jk} (a_j - \mu)^2 + (a_k - \mu)^2\\
&& + 2 (a_j -  \mu)(\mu - a_k)
\end{align*}
Since $(a_j - \mu)(\mu - a_k) \geq 0$, we have
\begin{align*}
\max_{jk} (a_j - a_k)^2 &\leq \max_j 4(a_j - \mu)^2\\
&\leq \sum_j 4(a_j - \mu)^2
\end{align*}
thus
\begin{align*}
\left\| \frac{\s(t)^\top}{\bm{\omega}(t)^\top} - \bar{x}\bm{1}^\top\right\|^2 	& \leq \frac{4\sum_{ij} \left(\P_{ji}(t) - \frac{1}{N}\sum_k\P_{ki}(t)\right)^2}{\min_{ij} \P_{ji}(t)^2} \|\bar{x}\bm{1}\|^2 \\
	&= \frac{4\|\J\P(t)\|^2_F}{\min_{ij} \P_{ji}(t)^2} \|\bar{x}\bm{1}\|^2
\end{align*}
with $\J = \bm{I} - \bm{11}^\top$.
We conclude the proof using the definition of Dobrushin's ergodicity coefficient
\begin{align*}
	\|\J\bm{A}\|^2 &=  \sum_{ij} \left(\bm{A}_{ij} - \frac{1}{N}\sum_k \bm{A}_{kj}\right)^2 \\
	&\leq \frac{1}{N^2}\sum_{ij} \left( \sum_k \vert \bm{A}_{ij} - \bm{A}_{kj} \vert \right)^2 \\
	&\leq \max_{ij}\left( \sum_k \vert \bm{A}_{ij} - \bm{A}_{kj} \vert \right)^2 = 4\mu_S^2(\bm{A})
\end{align*}
Thus, 
\begin{align*}
\left\| \frac{\s(t)^\top}{\bm{\omega}(t)^\top} - \bar{x}\bm{1}^\top\right\|^2  &\leq \frac{16\mu_S^2(\bm{A})}{\min_{ij}P_{ji}(t)^2} \|\bar{x}\bm{1}\|^2
\end{align*}
\end{IEEEproof}
\end{theorem}

As we can see from this proof, the convergence of the estimate to the average depends on the ratio between Dobrushin's ergodicity coefficient of the combined communication matrix and the minimum entry of the combined communication matrix. Dobrushin's ergodicity coefficient characterizes the speed at which the error decreases and depends on the choice for $\K$. Spectral bounds on convergence speed can be derived from ergodicity coefficients pretty easily, see \emph{e.g.},\cite{seneta2006non}.

The denominator is more challenging. The time $t$ at which the minimum entry of $\P(t) > 0$ characterizes the time at which the error is definite. Prior to that, some nodes did not receive any information, and although some error could be computed, such error bears no real meaning since these nodes did not even start to estimate the average. As such, this time is very important as it defines a minimum amount of exchanges that are needed for  all node estimates to be defined.

This minimum number of messages is closely related to the hitting time of a specific Markov chain, and is the main purpose of this paper. In the following sections, we give several upper bounds for this hitting time in the case of a gossip protocol that sends a single message at each time step.

\section{Hitting time generic bound} 

We consider the following sum-weight random gossip protocol defined by a graph $\GG$. The set of vertices of $V(\GG)$ are the nodes of the associated network, while the set of edges $E(\GG)$ are the available links between nodes.
We denote $N = \vert V(\GG) \vert$.
At each time step $t$, an edge $(s,r)$ is sampled from $E(\GG)$.

We assign a local variable $x_n(t)$ to each node $n$. At the beginning of the process, we arbitrarily chose one node $m$ such that its variable $x_m(0) = 1$ and set $\forall k\neq m, x_k(0) = 0$. Whenever a node sends a message, it contains a copy of its local variable $x_s(t)$, and when the message is received, the receiver $r$ updates its local variable with $x_r(t+1) = \min(1, x_r(t) + x_s(t))$. In other words, when a node receives a message, it sets its variable to $1$ if the message contains a $1$ or if it previously already had $1$.

Let us denote $R(t) = \sum_{k=1}^N x_k(t)$ the random variable that counts the number of nodes reached by the initial node at time $t$. We are interested in the number of exchanges needed such that all nodes are reached. For that purpose, we define the following random variable that counts the minimum time at which $k$ nodes were reached:
\begin{align}
	T_{k}=\inf\{t\geq0\mid R(t)=k\}
\end{align}
More specifically, we are interested in a probabilistic upper bound $T$ for $T_N$ the time at which all nodes are reached, or more formally $Pr[T_N \leq T] \geq \delta$.

Obtaining the expression of $p_k(i) = Pr[T_k = i]$ seems to be a challenging task and we rather focus on finding related concentration inequalities restrained to $p_N(i)$. For that matter, we need to find the moments of $T_N$, which we propose to obtain by using the probability generating function of $Pr[T_k = i]$ given in the following proposition:

\begin{proposition}
	The probability generating function $G_k(z)$ of $Pr[T_k = i]$ given by $G_k(z) = \sum_i p_k(i)z^i$ has the following expression:
	\begin{align}
	G_k(z) = z^{N-1}\prod_{k=1}^{N-1}\frac{\alpha_{k}}{1 - (1 - \alpha_{k})z}
	\end{align}
	with $\alpha_{k}=\alpha_{k}^{(i)}=Pr[R(i+1)=k+1\mid R(i)=k]$ is the probability that a new node is reached at time $i$ and depends on the structure of the graph.
	\begin{IEEEproof}
	Remark that we have the following marginalization on the number of nodes reached at time $i$:
	\begin{align*}
	p_k(i) &= \sum_{j<i}Pr[T_{k}=i\mid T_{k-1}=j]p_{k-1}(j)\\
	&= \sum_{j<i}\alpha_{k-1}(1-\alpha_{k-1})^{i-j-1}p_{k-1}(j)
	\end{align*}
	Using it in the definition of $G_k(z)$, we get
	\begin{align*}
	G_k(z) &= \sum_i p_k(i)z^i \\
	     &= \sum_i \sum_{j<i} \alpha_{k-1}(1-\alpha_{k-1})^{i-j-1}p_{k-1}(j)z^i\\
	     &= \sum_{u>0}\sum_v  \alpha_{k-1}(1-\alpha_{k-1})^{u-1}p_{k-1}(v)z^{u+v}\\
	     &= \sum_{u>0} \alpha_{k-1}(1-\alpha_{k-1})^{u-1}z^u\sum_vp_{k-1}(v)z^v\\
	     &= \sum_{u>0} \alpha_{k-1}(1-\alpha_{k-1})^{u-1}z^u G_{k-1}(z)\\
	     &= \alpha_{k-1}zG_{k-1}(z) \sum_{u>0}(1-\alpha_{k-1})^{u-1}z^{u-1}\\
	     &= \frac{\alpha_{k-1}z}{1 - (1 - \alpha_{k-1})z}G_{k-1}(z)
	\end{align*}
	Remark that since $R(0) = 1$, we have $G_1(z) = 1$ and consequently $G_k(1) = 1$ using the recursion. We can unroll the recursion to obtain $G_N(z)$ (all nodes reached):
	\begin{align*}
	G_N(z) = z^{N-1}\prod_{k=1}^{N-1}\frac{\alpha_{k}}{1 - (1 - \alpha_{k})z}
	\end{align*}
	\end{IEEEproof}
\end{proposition}
Using the probability generating function, we can now obtain the first and second order moments of $T_N$. The expectation of $T_N$ is given by the first derivative $G_N'(1)$ of $G_N(z)$ which leads to the following proposition
\begin{proposition}With $T_N$ and $\forall k, \alpha_k$ defined as previously, we have
\begin{align*}
	\EE[T_N] = \sum_{k=1}^{N-1}\frac{1}{\alpha_{k}}
\end{align*}
\begin{IEEEproof}
	We have 
	\begin{align*}
	\EE[T_n] = \sum_i p_n(i)i  = \sum_i p_n(i)i(1^i) = G_n'(1)
\end{align*}
Remark that 
\begin{align*}
	G_N'(z) = G_N(z)\frac{\partial \log(G_N)}{\partial z}(z)
\end{align*}
And in particular since $G(1) = 1$
\begin{align*}
	G_N'(1) &= \frac{\partial \log(G_N)}{\partial z}(1)\\
	&= \frac{\partial}{\partial z}\left[ log(z^{n-1}\prod_{k=1}^{n-1}\frac{\alpha_{k}}{1 - (1 - \alpha_{k})z})  \right]_{z=1}\\
	&= \left[\frac{n-1}{z} - \sum_{k=1}^{n-1} \frac{\alpha_{k} -1}{1 - (1-\alpha_{k})z}\right]_{z=1}\\
	&= n-1 - \sum_{k=1}^{n-1}\frac{\alpha_{k}-1}{\alpha_{k-1}} \\
	&= n-1 - \sum_{k=1}^{n-1}(1 - \frac{1}{\alpha_{k}}) = \sum_{k=1}^{n-1}\frac{1}{\alpha_{k}}
\end{align*}
\end{IEEEproof}
\end{proposition}
Using the same strategy, we can obtain the variance of $T_N$:
\begin{proposition}
	With $T_N$ and $\alpha_k$ defined as previously, we have
	\begin{align*}
	 \Var(T_N) =   \sum_{k=1}^{n-1}\frac{1}{\alpha_k^2} - \EE[T_n]
	\end{align*}
	\begin{IEEEproof}
	Remark that
\begin{align*}
\Var(T_N) &= \sum_i (i-\mu)^2p_N(i)\\
&= \sum_i i^2p_N(i) - \left(\sum_i ip_N(i)\right)^2\\
&= \sum_i i(i-1)p_N(i) + \sum_i ip_N(i)\\
&\qquad\qquad\qquad\quad\qquad - \left(\sum_i ip_N(i)\right)^2\\
&= G_N''(1) + G_N'(1) - (G_N'(1))^2\\
&= \EE[T_N] + G_N''(1) - (G_N'(1))^2
\end{align*}
Using the following property:
\begin{align*}
	\frac{\partial}{\partial z}\left[ \frac{G_N'}{G_N} \right]_{z=1} &= \left[ \frac{G_N''G_N - G_N'^2}{G^2}\right]_{z=1} \\
	&= G_N''(1) - G_N'(1)^2\\
	&= \left.\frac{\partial^2}{\partial z^2} \log(G_N) \right]_{z=1}\\
	&= \left.\frac{1-N}{z^2} + \sum_{k=1}^{N-1} \frac{(1-\alpha_k)^2}{(1 - (1-\alpha_k)z)^2} \right]_{z=1}\\
	&= 1-N + \sum_{k=1}^{N-1} \frac{(1-\alpha_k)^2}{\alpha_k^2}\\
	&= 1-N + \sum_{k=1}^{N-1} \left( 1 - \frac{1}{\alpha_k} \right)^2\\
	&=  \sum_{k=1}^{n-1} \frac{1}{\alpha_k^2} - 2 \sum_{k=1}^{N-1}\frac{1}{\alpha_k}\\
	&= \sum_{k=1}^{N-1} \frac{1}{\alpha_k^2} - 2\EE[T_N] 
\end{align*}
By replacing $G_N''(1) - G_N'(1)^2$ in the expression of $\Var(T_N)$ we obtain the result.
	\end{IEEEproof}
\end{proposition}
Using these moment, we can now give a probabilistic upper bound for $T_N$ using Chebyshev inequality.
\begin{theorem}
	With probability at most $\delta$, we have 
	\begin{align*}
		T_n \geq \sum_{k=1}^{N-1}\frac{1}{\alpha_{k}} + \sqrt{\frac{\sum_{k=1}^{N-1}\frac{1}{\alpha_{k}^2} -\sum_{k=1}^{N-1}\frac{1}{\alpha_{k}} }{\delta}}
	\end{align*}
	\begin{IEEEproof}
	We use Chebyshev inequality which states $Pr[ \vert T_N - \EE[T_N] \vert \geq k\sqrt{\Var(T_N)} ]  \leq 1/k^2 $. Setting the right hand side to $\delta$ leads to $k = 1/\sqrt{\delta}$.
	Replacing $\EE[T_N]$ and $\Var(T_N)$ in this expression and considering only the positive part gives the desired result.
	\end{IEEEproof}
\end{theorem}

In practice however, this theorem is difficult to use, because it requires all transition probabilities $\alpha_k$ to be known, which depend on the topology of the graph. These values can be computed for some special cases but are generally inaccessible. In the following we propose a usable expression for the simple case of fully connected graphs.

\section{Special case of fully connected graphs}

In the case of fully connected graphs, the probability of reaching a new node is simply the probability that the sender was picked from reached nodes and that the receiver was picked from non-reached nodes
\begin{align*}
	\alpha_k = Pr[ R(t+1) = k+1 | R(t) = k ] = \frac{k}{N}\frac{N-k}{N-1}
\end{align*}

This allows us to bound both the mean and the variance of $T_N$
\begin{proposition}
	For a fully connected graph, we have 
	\begin{align*}
	\EE[T_N] \leq 2(N-1)\log(N+1)
	\end{align*}
	\begin{IEEEproof}
	We have 
	\begin{align*}
	\EE[T_N] &= \sum_{k=1}^{N-1}\frac{1}{\alpha_k} = \sum_{k=1}^{N-1} \frac{N(N-1)}{k(N-k)} \\
	&= (N-1)\sum_{k=1}^{N-1} \frac{N-k+k}{k(N-k)} \\
	&= (N-1) \sum_{k=1}^{N-1}\left(\frac{1}{k} + \frac{1}{N-k}\right)\\
	&= 2(N-1) \sum_{k=1}^{N-1}\frac{1}{k}
	\end{align*}
	We recognize the harmonic number which can be bounded by $\log(N+1)$ and leads to the result.
	\end{IEEEproof}
\end{proposition}

\begin{proposition}
	For a fully connected graph, we have
	\begin{align*}
	\Var(T_N) \leq \frac{\pi^2(N-1)^2}{3}
	\end{align*}
	\begin{IEEEproof}
	We have 
	\begin{align*}
	\Var(T_N) &= \sum_{k=1}^{N-1}\frac{1}{\alpha_k^2} - \EE[T_N] = \sum_{k=1}^{N-1} \frac{N^2(N-1)^2}{k^2(N-k)^2} - \EE[T_N]\\
	&\leq (N-1)^2 \sum_{k=1}^{N-1} \frac{N^2}{k^2(N-k)^2}  \\
	&= (N-1)^2 \sum_{k=1}^{N-1} \frac{N^2-2k+k^2+2k-k^2}{k^2(N-k)^2} \\
	&= (N-1)^2 \sum_{k=1}^{N-1}\left( \frac{1}{k^2}- \frac{1}{(N-k)^2}\right. \left. +\frac{2}{k(N-k)^2} \right)  \\
	&\leq (N-1)^2 \sum_{k=1}^{N-1} \frac{2}{(N-k)^2}  \\
	&\leq \frac{\pi^2(N-1)^2}{3}
	\end{align*}
	\end{IEEEproof}
\end{proposition}
Using these results, we can give a proper bound on the total diffusion time in the case of fully connected graphs.
\begin{theorem}
	For a fully connected graph, we have with probability at most $\delta$
	\begin{align*}
	T_N \geq 2(N-1)\log(N+1) + \frac{\pi(N-1)}{3\sqrt{\delta}}
	\end{align*}
\end{theorem}
The main appeal of this result is that it show a total diffusion can be reached with a logarithmic number of messages sent per node. Indeed, if we set each node to send $T_N/N$ messages, which does not require more coordination than just knowing the total of nodes in the graph, we are assured to reach full diffusion with probability $1-\delta$. 

Remark also that this result can be compared to the ``telephone call problem'' discussed in~\cite{frieze1985shortest}, where at each step all reached nodes send a message to a random destination. In that case, $T_N = \mathcal{O}(\log N)$ in probability, but the total number of messages sent is difficult to obtain since it depends on the number of reached nodes at each time step $t$. An approximation would be that a maximum of $N$ messages are sent at each time step $t$ and thus we recover $\mathcal{O}(N\log N)$.
\section{Case of bottlenecked graphs}
\label{sec:cheeger}
In the case of non-fully connected graphs, total diffusion time can grow slowly if the graph contains bottlenecks. These bottlenecks lower the probability of reaching new nodes in one side of the bottleneck when starting from the other side. A measure of the ``bottleneckedness'' of a graph $\GG$ is given by Cheeger's constant $\Phi_C(\GG)$:
\begin{align*}
	\Phi_C(\GG) = \min_{S \subset V(\GG)}\frac{\vert \partial S \vert}{\min(\vert S \vert, \vert V(\GG) \setminus S \vert)}
\end{align*}
with $\partial S = \{ (s,r) \in E(\GG) \vert s \in S, r \in V(\GG)  \setminus S \}$.

The relative volumes of these subgraphs can be expressed in terms of probability of drawing a pair of linked nodes such that the first is in $S$ and the second is not:
\begin{align*}
	\Pr[(r,s)\in E(\GG) \vert s \in S, r \in V(\GG) \setminus S]  = \frac{\vert \partial S \vert }{\vert E(\GG) \vert}
\end{align*}

We use this notation to bound the $\alpha_k$ in the following proposition:
\begin{proposition}
	For any graph $\GG$, we have
	\begin{align*}
	\alpha_k \geq \Phi_C(\GG) \frac{\min(k, N-k)}{\vert E(\GG) \vert}
	\end{align*}
	\begin{IEEEproof}
	Let us denote $S_k = \{ s \in V(\GG) \vert x_s(t) = 1 \}$ the set of reached nodes. Recall that
	\begin{align*}
	\vert \partial S_k \vert &= \Pr[s \in S_k, r \in V(\GG) \setminus S_k] {\vert E(\GG) \vert} \\
	&= \alpha_k \vert E(\GG) \vert
	\end{align*}
	Using its definition, we can bound Cheeger constant using our special subset $S_k$:
	\begin{align*}
	\Phi_C(\GG) &\leq \frac{\vert \partial S_k \vert}{\min(\vert S_k\vert, \vert V(\GG) \setminus S_k \vert)} \\
	&= \frac{\alpha_k \vert E(\GG) \vert}{\min(k, N-k)} 
	\end{align*}
	\end{IEEEproof}
\end{proposition}

Having a lower bound for the $\alpha_k$ allows us to bound the expectation of $T_N$ with the following proposition:
\begin{proposition}
	Let $\Phi_C(\GG)$ be the Cheeger constant of $\GG$, then
	\begin{align*}
	\EE[T_N] \leq \frac{2 \vert E(\GG) \vert}{\Phi_C(\GG)} \log\left(\frac{N}{2}+1\right)
	\end{align*}
	\begin{IEEEproof}
	Recall that
	\begin{align*}
	\EE[T_N] &= \sum_{k=1}^{N-1} \frac{1}{\alpha_k} \leq  \frac{\vert E(\GG) \vert}{\Phi_C(\GG)} \sum_{k=1}^{N-1} \frac{1}{\min(k, N-k)} \\ 
	& \leq 2\frac{\vert E(\GG) \vert}{\Phi_C(\GG)}\log\left(\frac{N}{2}+1\right)
	\end{align*}
	\end{IEEEproof}
\end{proposition}

Similarly, we can obtain a bound on the variance of $T_N$
\begin{proposition}
	Let $\Phi_C(\GG)$ be the Cheeger constant of $\GG$, then
	\begin{align*}
	\Var(T_N) \leq \frac{\pi^2\vert E(\GG)\vert^2}{3\Phi_C(\GG)^2}
	\end{align*}
	\begin{IEEEproof}
	We have
	\begin{align*}
	\Var(T_N) &= \sum_{k=1}^{N-1}\frac{1}{\alpha_k^2} - \EE[T_N] \\
	&\leq \frac{\vert E(\GG)\vert^2}{\Phi_C(\GG)^2}\sum_{k=1}^{N-1} \frac{1}{\min(k, N-k)^2}  \\
	&\leq \frac{\pi^2\vert E(\GG)\vert^2}{3\Phi_C(\GG)^2}
	\end{align*}
	\end{IEEEproof}
\end{proposition}

With bounds for $\EE[T_N]$ and $\Var(T_N)$, we can prove the final theorem of this paper which bounds the number of messages needed to achieve full diffusion in terms of conductance of the associated graph:
\begin{theorem}
	Let $\Phi_C(\GG)$ be the Cheeger constant of $\GG$, then with probability at most $\delta$
	\begin{align*}
	T_N \geq \frac{2 \vert E(\GG) \vert}{\Phi_C(\GG)} \log\left(\frac{N}{2}+1\right) + \frac{\pi\vert E(\GG) \vert}{\Phi_C(\GG)\sqrt{3\delta}}
	\end{align*}
	\begin{IEEEproof}
	Application of Chebyshev inequality with probability $\delta$ using the previous bounds on $\EE[T_N]$ and $\Var(T_N)$.
	\end{IEEEproof}
\end{theorem}
Contrarily to the case of fully connected graphs, the bound is no longer always dominated by $N$ and grows in $\vert E(\GG)\vert$.

Remark also that Cheeger's inequality links $\Phi_C(\GG)$ with the second smallest eigenvalue of the graph and can thus be used in place of $\Phi_C(\GG)$. This leads to the following corollary:
\begin{corollary}
	Let $\lambda_2$ be the second eigenvalue of $\GG$, then with probability at most $\delta$
	\begin{align*}
	T_N \geq \frac{2 \vert E(\GG) \vert}{\sqrt{2\lambda_2}} \log\left(\frac{N}{2}+1\right) + \frac{\pi\vert E(\GG) \vert}{\sqrt{6\lambda_2\delta}}
	\end{align*}
	\begin{IEEEproof}
	Recall and apply Cheeger's inequality~\cite{kwok13acm}:
	$\Phi_c(\GG) \leq \sqrt{2\lambda_2}$
\end{IEEEproof}
\end{corollary}
 However, computing this eigenvalue is often impractical since it requires the adjacency matrix of the complete graph, which is often unavailable in decentralized contexts. In some cases (\textit{e.g.}, Barab{\'a}si-Albert graphs), $\lambda_2$ can nonetheless be bounded without knowing the exact graph.

\section{Simulation}

In this section we show simulations of $T_N$ for different graphs and compare them to our proposed bounds. In each example, we made 50 random realizations of the gossip protocol and record the time $T_N$ at which all nodes were reached.

\begin{figure}[t!]
\centering
	\includegraphics[width=\columnwidth]{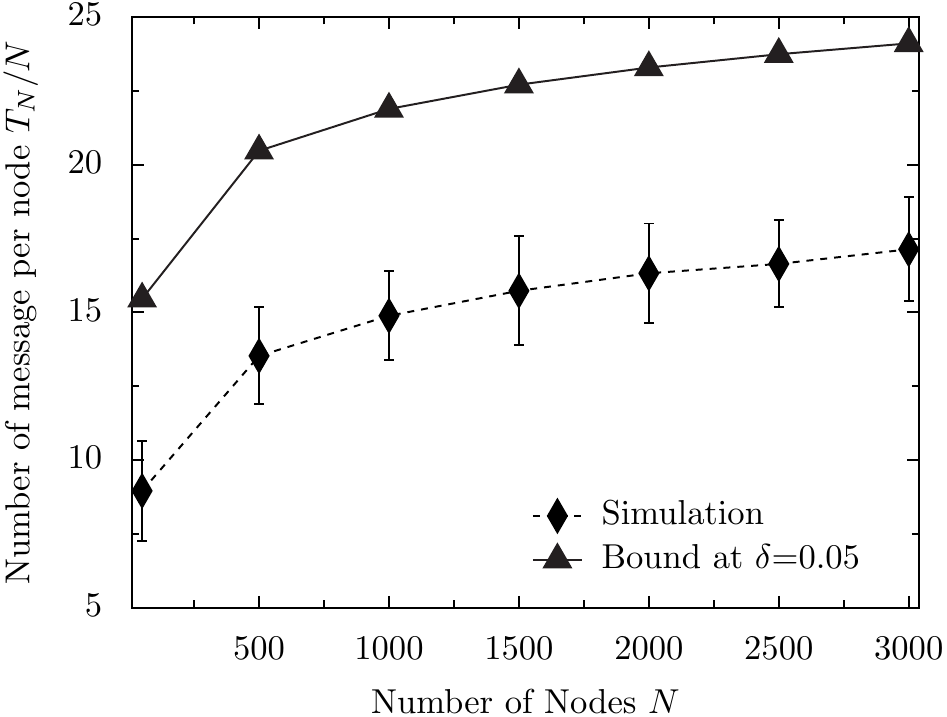}
	\caption{Empirical values for $T_N$ and corresponding bound for a fully connected network with varying size $N$.}
	\label{fig:full}
\end{figure}

We first simulate a fully connected network of size $N$ and plot the mean number of messages per node $T_N/N$ and its standard deviation on figure~\ref{fig:full}. We also show the corresponding bound for $\delta=0.05$.
As we can see, this number increases logarithmically with a constant standard deviation as predicted by our bound.
Even for large network of over $1000$ nodes, the gap between the predicted number of messages and the empirical number of messages only differ by less than 5.

\begin{figure}[t!]
\centering
	\includegraphics[width=\columnwidth]{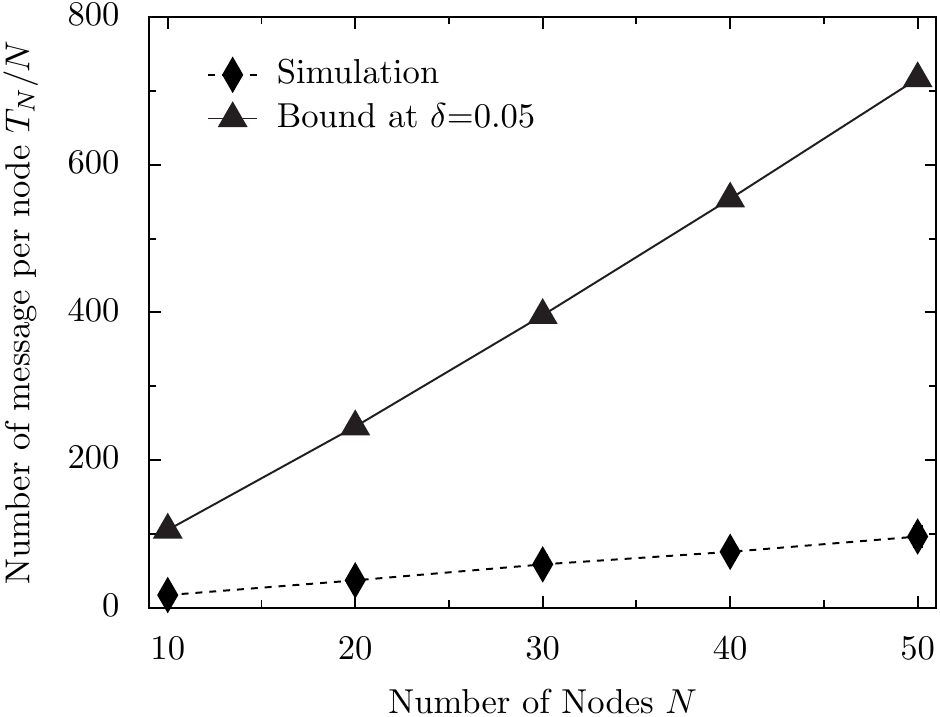}	
	\caption{Empirical values for $T_N$ and corresponding bound for a chain network with varying size $N$.}
	\label{fig:chain}
\end{figure}

Next, we simulate a chain network of size $N$. The number of edges in the associate graph is $N-1$ while the Cheeger constant is $2/N$. The bottleneck occurs in the middle of the chain, since one edge links the 2 halves of the network.
We plot the empirical $T_N$ and its standard deviation on figure~\ref{fig:chain}.
As we can see, our bound using Cheeger constant is pretty far from the empirical value but captures the behavior of the system. Indeed, in a chain, the bottleneck on the probability $\alpha_k$ is not quadratic but linear thanks to a linear number of edges and every link is a kind of bottleneck.

\begin{figure}[t!]
\centering
	\includegraphics[width=\columnwidth]{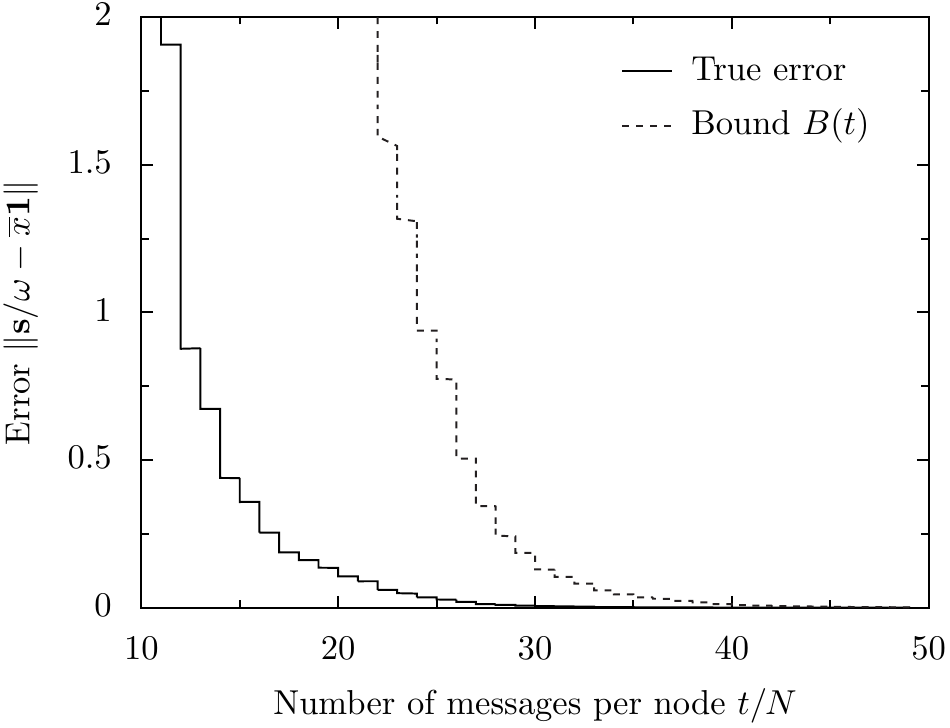}
	\caption{Evolution of the true error with the bound used to prove the asymptotic convergence.}
	\label{fig:dob}
\end{figure}

finally, we show on figure~\ref{fig:dob} the evolution of the error $\left\| \frac{\s^\top}{\bm{\omega}^\top} - \bar{x}\bm{1}^\top\right\|^2$ compared to the first upper bound we use in the proof of theorem~\ref{th:random} given by 
\begin{align*}
	B(t) = \left\|\frac{\max_{ji}\mathbf{P}(t) - \min_{ki}\mathbf{P}(t)}{\min_j\mathbf{P}_i(t)}\right\|
\end{align*}, for a fully connected graph of size $N=100$. We initialize all estimates to 0 with the exception of the first node having $\s_0(t) = N$. All weights are initialized by 1.
Recall that for such graph, our hitting time bound states that such error is finite with probability $95\% $ only after $t/N = 16$ messages per node. Remark that the error is finite before that. While $B(t)$ is not usable before $t/N = 25$ mainly due to the very small values of $\P(t)$, it still gives a reasonable approximation at the later stages. In practice however, computing $B(t)$ would require the knowledge of $\P(t)$ which is not possible in a decentralized setup.

\section{Conclusion}

In this paper, we investigate non-asymptotic bounds for randomized sum-weights gossip protocols that are essential for algorithm that need to select a termination time. We show that such times are depending on two factors, namely the speed at which the error decreases and the time at which all nodes are sufficiently confident in their estimate. The first term has already been studied and we show it is related to Dobrushin's ergodicity coefficient that depends on the choice of distribution from which the communication matrices are sampled. We show that the second term is related to the hitting time of the Markov chain that counts the number of nodes that have received at least one message. We obtain non-asymptotic bounds to the number of messages required such that all nodes have been hit. We extends these results to the case of fully connected graph for which we obtain a bound that matches experimental simulation. We also consider the case of bottlenecked graphs, where we show the hitting time depends on Cheeger constant which measures the ``bottleneckedness'' of the considered graph.

\bibliographystyle{IEEEtran}
\bibliography{mybibfile}

\end{document}